\documentclass[runningheads]{llncs}
\usepackage{graphicx}
\usepackage{multirow}
\usepackage{cite}
\usepackage{xcolor}
\usepackage{float}
\floatstyle{plaintop}
\restylefloat{table}
\begin{document}
\title{Deep Reinforcement Learning solution for pickup and delivery routing problems with time window and capacity constraints}
\titlerunning{CPDPTW with deep RL}
%
\author{Andrew Soroka\inst{1} \and
Alex Meshcheryakov\inst{2} \and
Sergey Gerasimov\inst{3}}
\authorrunning{A. Soroka et al.}
%
\institute{Moscow State University, Moscow, Russian Federation \email{andrew.soroka@student.msu.ru} \and Space Research Institute of RAS, Moscow, Russia \email{ mesch@cosmos.ru} \and Moscow State University, Moscow, Russian Federation \email{gerasimov@cs.msu.ru}}
\maketitle              
\begin{abstract}
\textcolor{black}{The task of constructing vehicles optimal routes for pickup and delivery} of goods is one of most promising tasks in the context of global urban population growth. Although this kind of problems with small size can be solved by various classical approaches, a fast (or real-time) route optimizer under the constraints of the real world (such as capacity and time windows constraints) for medium-large size problems still remains a highly challenging task. In this work we, for the first time, successfully applied a deep Reinforcing Learning approach \textcolor{black}{(modified JAMPR model)} to solve Pickup
and Delivery problem with Capacity and Time Window constraints (CPDPTW). We obtained a robust model that gives a fast optimal solution for problems of small and medium size, and gives fast suboptimal solution for problems of larger ($>200$) size.

\keywords{vehicle routing problems  \and deep learning \and reinforcement learning.}
\end{abstract}
\section{Introduction}
The Vehicle Routing Problem (VRP) is a combinatorial optimization and integer programming problem \cite{wolsey1999integer}. It answers the question, ''What is the optimal set of routes for a fleet of traveling vehicles to deliver a given set of customers?''. VRP generalizes the well-known traveling salesman problem (TSP) \cite{parragh2008survey} and both are NP-hard. 

In the real world, there are many additional restrictions dictated by the need to fulfill real world business requests. So, customers do not always only receive the goods, sometimes they want to hand over the goods to the depot, or deliver them to another client, --- the pickup and delivery (PDP) restriction appears. The client's cargo volume is actually non-zero, and there are restrictions on the maximum volume (Capacity) of vehicles and the need to handle it. Availability of clients during certain time slots adds another kind of real world constraints to pay attention to --- a Time Windows. All these restrictions add up to the Pickup and Delivery problem with Capacity and Time Window constraints (CPDPTW).

The CPDPTW problem has many practical applications: courier delivery, taxi operation, logistics of goods between warehouses and points of sale. There are tools that allows one to get a suboptimal solution for the classical problem without constraints and with limited size (for example, Google OR-Tools \cite{perron2011operations}, heuristics \cite{braysy2005vehicle, clarke1964scheduling}, deep neural networks \cite{nazari2018reinforcement}). But there is still no proposed solution for a large-dimension routing problem with all important real world constraints.

A well-known issue with existing approaches is the difficulty of handling new real-world constraints. For example, the implementation of HGS \cite{vidal2022hybrid, vidal2012hybrid} heuristic only supports solving the classical CVRP problem. From the other hand, another popular heuristic, OR-Tools\cite{perron2011operations}, supports many realistic constraints but is not always able to give a feasible solution even for small problems (see e.g. Table 1 in \cite{kool2018attention}). Despite the fact that powerful heuristic solvers (like LKH-3 \cite{lin1973effective} or HGS \cite{vidal2022hybrid, vidal2012hybrid}) finally can find a good solution for large ($>$1000) problems, they suffer from a substantial manual labor in the creation of successful heuristics and huge computational load due to long iterative calculations --- the consequence is a lack of model flexibility. For example, LKH-3 takes more than an hour to solve a CVRP task instance of size 2000, which is inappropriate for many applications such as large courier or municipal services \cite{li2021learning}.

The motivation of our work is to create a fast neural solver capable of handling high-dimensional problems with the limitations inherent in the real world: time windows, limited vehicle volume, multi-depot. To solve this task, we considered reinforcement learning algorithms, since the strategies for solving the VRP optimization problem can be accurately parameterized using neural networks. We aim to explore the applicability of reinforcement learning algorithms based on the JAMPR model to solve CPDPTW problems and compare them to baseline solution provided by OR-Tools heuristic. 

\section{Related work}
\subsection{Classical approaches}
There are two classical approaches to solve routing problems: mixed-integer programming (MIP) and heuristics. MIP is a class of mathematical programming problems in which the problem domain is specified by inequalities over real and/or integer variables. A very important special case is mixed-integer linear programming, which further restricts the problem to linear inequalities and linear objective function \cite{wolsey1999integer}. The two main algorithms, used in MIP for solve routing problems is cutting plane methods \cite{dantzig1954solution} and branch and bound algorithms \cite{little1963algorithm}. Despite the fact, that MIP solutions are precise, integer programming requires too much time to generate solution even for small dimension problems.

Heuristic algorithms are used most often for such problems and can be roughly divided into constructive heuristics and metaheuristics. Constructive heuristics are procedures that construct a correct solution to a VRP problem, usually in polynomial time in the size of the input data, and with no guarantees in terms of the quality of the solution. The best-known design heuristic is the nearest neighbor algorithm, which builds vehicle routes step by step, choosing the nearest available location at each step, starting at the depot \cite{clarke1964scheduling}. While design heuristics generally guarantee a fast solution that satisfies constraints, there is no guarantee that the solution is optimal. That is why the most advanced heuristics use a search procedure.

Metaheuristics are procedures that repeatedly use simple rules or simpler heuristics to construct an optimal route. The simplest and most common example of a metaheuristic is a local search — a family of search procedures that have the following common features: 
\begin{enumerate}
    \item start the search with a single feasible solution (usually generated using some constructive heuristic)
    \item at each iteration create a neighborhood of the existing solution - a set of valid solutions, usually of polynomial size and consisting of solutions "similar"\ to the current solution
    \item for the neighborhood of the current solution, one solution is chosen and adopted (raised to the current solution) in accordance with some rule, which can be either deterministic (adopted if it is better than the current one) and stochastic (accepted if it is better, otherwise accepted with some probability) \cite{braysy2005vehicle}.
\end{enumerate}

These also include local search, genetic algorithms and ant colony methods. For example, move, exchange, and 2-opt are well-known heuristics for traveling salesman problems and vehicle routing. The most common LKH-3 VRP solver uses the Lin-Kernighan heuristic (replacing sub-route pairs to create new routes) as a basis, while the HGS CVRP solver uses a hybrid genetic algorithm and a local search procedure to achieve competitive results for problems up to 1000 in size. Additional approach, often used in real-world applications, is the open source solution from Google OR-Tools \cite{perron2011operations}. The concept of the algorithm is to start with an initial possible solution, which is generated using relatively simple heuristics such as the cheapest path arc (PCA) heuristic. The original solution is then iteratively improved by applying a set of change statements to the current solution \cite{braysy2005vehicle}. For routing and vehicle scheduling problems, the most suitable class of improvement operators are the so-called edge exchange algorithms. Among the most common enhancement operators used in OR-Tools for VRP and PDP are Two-opt \cite{lin1973effective}, OR-opt \cite{or1976traveling}, Relocate, Exchange, and Cross \cite{savelsbergh1992vehicle}. We use Google OR-Tools as a baseline for our research as its main tool used in real world tasks.

\subsection{DL and RL algorithms}
The first deep learning model for sequential VRP solution was proposed by Nazari et al. \cite{nazari2018reinforcement} who adapted the Pointer Network (PtrNet) Vinyals et al. \cite{vinyals2015pointer} for working with CVRP. Nazari et al. \cite{nazari2018reinforcement} completely abandoned the original part of the RNN encoder model and replaced it with a layer of linear embedding with common parameters. A more recent AM algorithm by Kool et al. \cite{kool2018attention} replaced this architecture with an adapted transformer model using self-attention \cite{vaswani2017attention}. A direct improvement to this model is the JAMPR approach by Falkner et. al \cite{falkner2020learning}, where the authors added additional linear embedding networks for the current path and position of the trucks. This add-on allowed the algorithm to successfully solve CVRP-TW problems. Chen and Tian \cite{chen2019learning} propose an RL-based improvement approach that iteratively selects a region on a graph and then selects and applies established local heuristics. This approach has been further improved by the destruction operator introduced by Lu et al. \cite{lu2019learning}. The latest attempt to use deep learning to partition a set of points into subproblems and solve it with a black box solver is proposed by Li et al. \cite{li2021learning}. The authors proposed two versions of the supervised approach: regression prediction of a possible improvement in the final cost and classification into the best subtask. By reducing the dimension and using classical metaheuristic approaches in each subproblem, the authors managed to show good results on problems of high dimension (more than 1000 points). To the best of our knowledge, there are currently no studied deep learning approaches for high-dimensional CPDPTW problems, so our work may be of interest as a baseline for solving real-world routing problems.

\section{Model}

\textcolor{black}{We used the JAMPR \cite{falkner2020learning} model as the main algorithm. The original JAMPR model is a modification of the Attention Model (AM) \cite{kool2018attention} of Kool et al., which uses an encoder-decoder architecture with self-attention \cite{vaswani2017attention}. Both models treat the route optimization problem as a sequential decision problem, which is modeled as a Markov decision process and solved using reinforcement learning. The solution to the problem is built piece by piece by creating routes one node at a time. The current decision, route and unvisited nodes are interpreted as state, and the index of all unvisited nodes that can be added to the current route as actions.}

\begin{figure}[!h]
\centering
\includegraphics[width=0.6\textwidth]{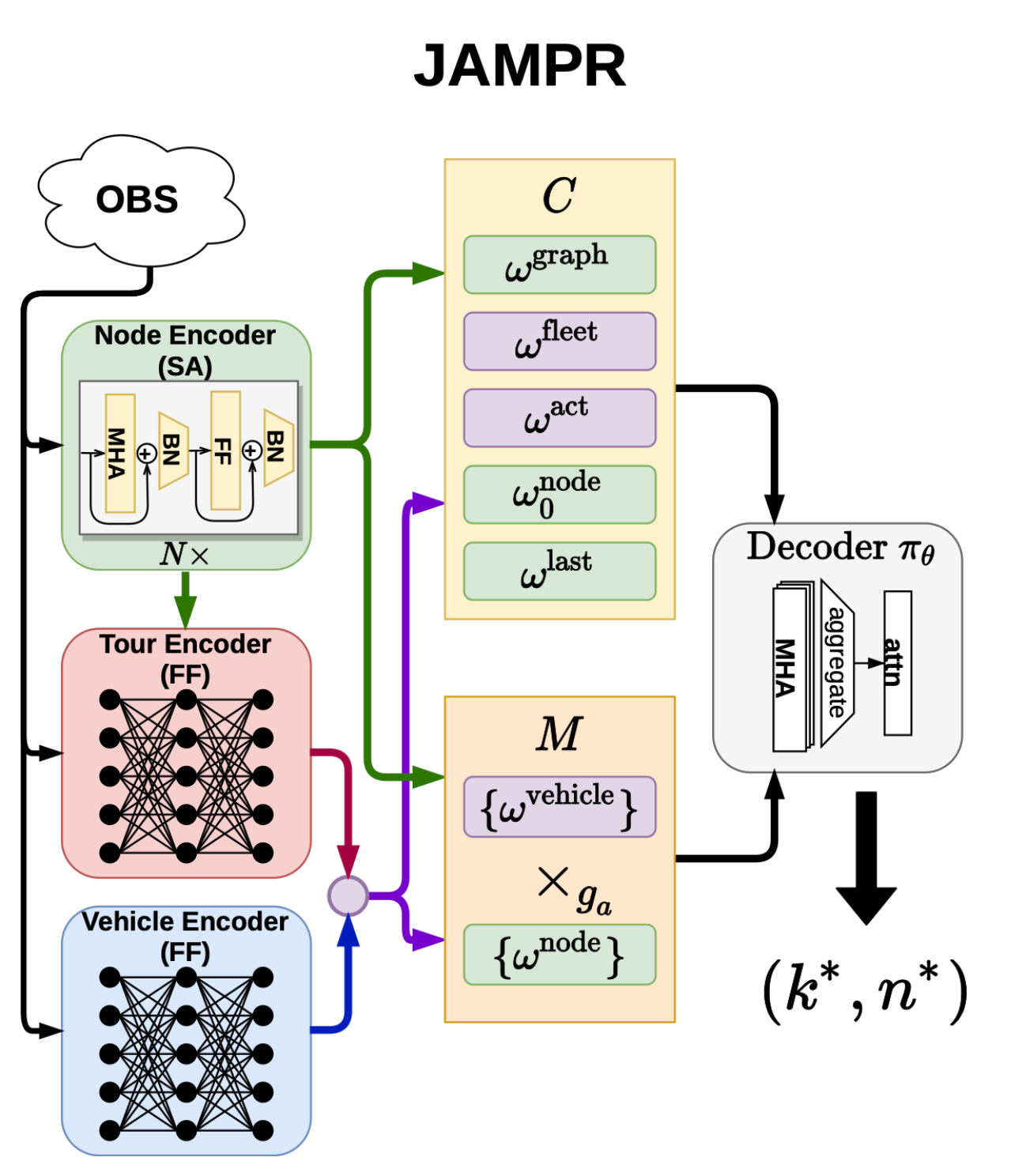}
\caption{An original image of JAMPR architecture. Source: Figure 1 from \cite{falkner2020learning}}
\label{fig:jampr}
\end{figure}

\textcolor{black}{First, the encoder receives the node features $x_i$ of each node $i$ (coordinates, demands, time windows, etc.) and encodes them into a hidden embedding vector $\tilde{x_i} \in R^{d_{emb}}$ with dimension $d_{emb}$. The decoder model then computes an attention query for each $\tilde{x_i}$ w.r.t. a specific context $C^{(t)}$ at decoding step $t$ to obtain estimates for all nodes that can be added to the current route. Here, the context includes the implicit embedding of the problem graph and additional information about the problem, such as the depot node index, the last node added to the current route, and the remaining bandwidth. The resulting scores are then either used in a greedy selection procedure, i.e. the node with the highest score is always selected, or softmax is used to transform it into a distribution that is used for sampling. In general, the encoder-decoder model is a policy $\pi(i^{(t+1)} \vert C^{(t)},x;\theta)$ with trainable parameters $\theta$. You can see author's visualization of architecture from original paper \cite{falkner2020learning} on figure \ref{fig:jampr}.}

\textcolor{black}{JAMPR \cite{falkner2020learning} has extended the standard AM \cite{kool2018attention}, which was primarily designed for TSP and CVRP, with additional encoders for routes and vehicles to enrich the context for the VRPTW solution. To do this, JAMPR creates a hidden embedding for each constructed route $r \in R$ by introducing node embeddings $\tilde{x_i},$ $i \in r$ and vehicle characteristics $\phi_r$ (residual capacity, current node, current time, etc.) of the corresponding route $r$ into additional fully connected neural networks that aggregate output and combine it with context. Since the extended context created in this way is a more complete representation of the state, it allows multiple routes to be built in parallel, which has proven necessary to successfully solve highly restrictive problems such as VRPTW. The number of such simultaneously planned routes is fixed by a constant $\kappa$. This leads to a new extended action space of currently active routes $R^{(t)}_\kappa$ and available nodes equal to $A = \{(r, i) \in feasible(R^{(t)}_{\kappa} \times N)\}$, where $N$ is the number of clients and $feasible()$ is a function that selects only those nodes that can be added to routes without violating any restrictions.}

\textcolor{black}{We have modified $feasible()$ function to support the PDP restriction. We added an additional mask that limits the available customers for each truck according to the specified delivery order, similar to the masks of time windows and visited customers. In real-life scenarios, being able to visit all customers may not be feasible, but giving up a solution, even a partial one, is impractical. We decided not to consider the HARD statement of the problem (the absence of a solution when at least one client is missed). To support SOFT setting, the final cost of the route was changed: the final result is a linear combination of the distance traveled and the number of missed clients with coefficients of 13 and 10, respectively, both during training and during inference. These values were chosen empirically.}

\section{Data} \label{data}
For CPDPTW, we select suitable task instances from a distribution based on the R201 statistic, Solomon's well-known reference set \cite{solomon1987algorithms}. Truck volumes are given as $Q_{50} = 750$, $Q_{100} = 1000$, $Q_{200} = 3500$, $Q_{400} = 6000$, $Q_{1000} = 12000$ for problems of sizes 50, 100, 200, 400 and 1000, respectively. The total time horizon is $[a_0, b_0]$, where $a_0 = 0$ for all examples, while the right boundary $b_0$ varies depending on the task size: 1000 for 50 and 100 points, 2000 for 200 and 400 points, 4000 for 1000 points. While service duration $h_i$ is uniformly set to 10. To simulate pdp, we randomly select from 3 to 7 points, marking them as depots (pick up points), the rest of the points are evenly distributed among the previously selected depots and marked by customers (delivery points).

\section{Experiments}
To analyze the algorithm, we divided the CPDPTW problems by size into three main classes:
\begin{itemize}
    \item low-dimensional problems (50 points)
    \item medium size problems (100, 200 points)
    \item high-dimensional problems (400, 1000 points)
\end{itemize}

For each task size, we trained the JAMPR model on dynamically generated task instances from the distribution described in the \ref{data} section. The model was trained using the early stopping criterion: no decrease in the cost of the path over 10 optimization steps. In the table \ref{table} you can see that for almost all dimensions considered, the time spent on training models is measured in days. We note that despite the results obtained in the study, the model requires an impressive training time.

\begin{table}[h]
\centering
\begin{tabular}{|l|l||l|l|}
\hline
                        & Problem size                & Epoch & Training time (hours) \\ \hline
\multirow{2}{*}{Small}  & \multirow{2}{*}{50} & 10    & 3            \\ 
                        &                     & 723   & 163          \\ \hline
\multirow{2}{*}{Medium} & 100                 & 23    & 47           \\ 
                        & 200                 & 21    & 93           \\ \hline 
\multirow{2}{*}{Large}  & 400                 & 25    & 188          \\ 
                        & 1000                & 10    & 336          \\ \hline
\end{tabular}
\caption{Table with the time spent on training all the presented models for the CPDPTW task.}
\label{table}
\end{table}

The trained model was tested on 100 instances of problems of each dimension, except for the size of 1000 points, for which 10 pre-generated test cases were taken.

All tests were performed on a server with an NVIDIA Tesla A40 GPU and an Intel(R) Xeon(R) Gold 6226R CPU @ 2.90GHz.

\subsection{Problems of small size}
To test the model on low-dimensional problems, we trained JAMPR for about 7 days. We selected two versions of network states: 10 training epochs (approximately 3 hours) and 723 epochs (approximately 7 days) to show how the predictive ability of the network changes with increasing training time. For the gap chart, we calculated the percentage deviation of the models at each point compared to the final result shown by OR-Tools for 100 seconds of optimization. The trained models were tested on 100 task instances, the final metric is the average value among all solved task instances.

\begin{figure}[!h]
\centering
\includegraphics[width=0.9\textwidth]{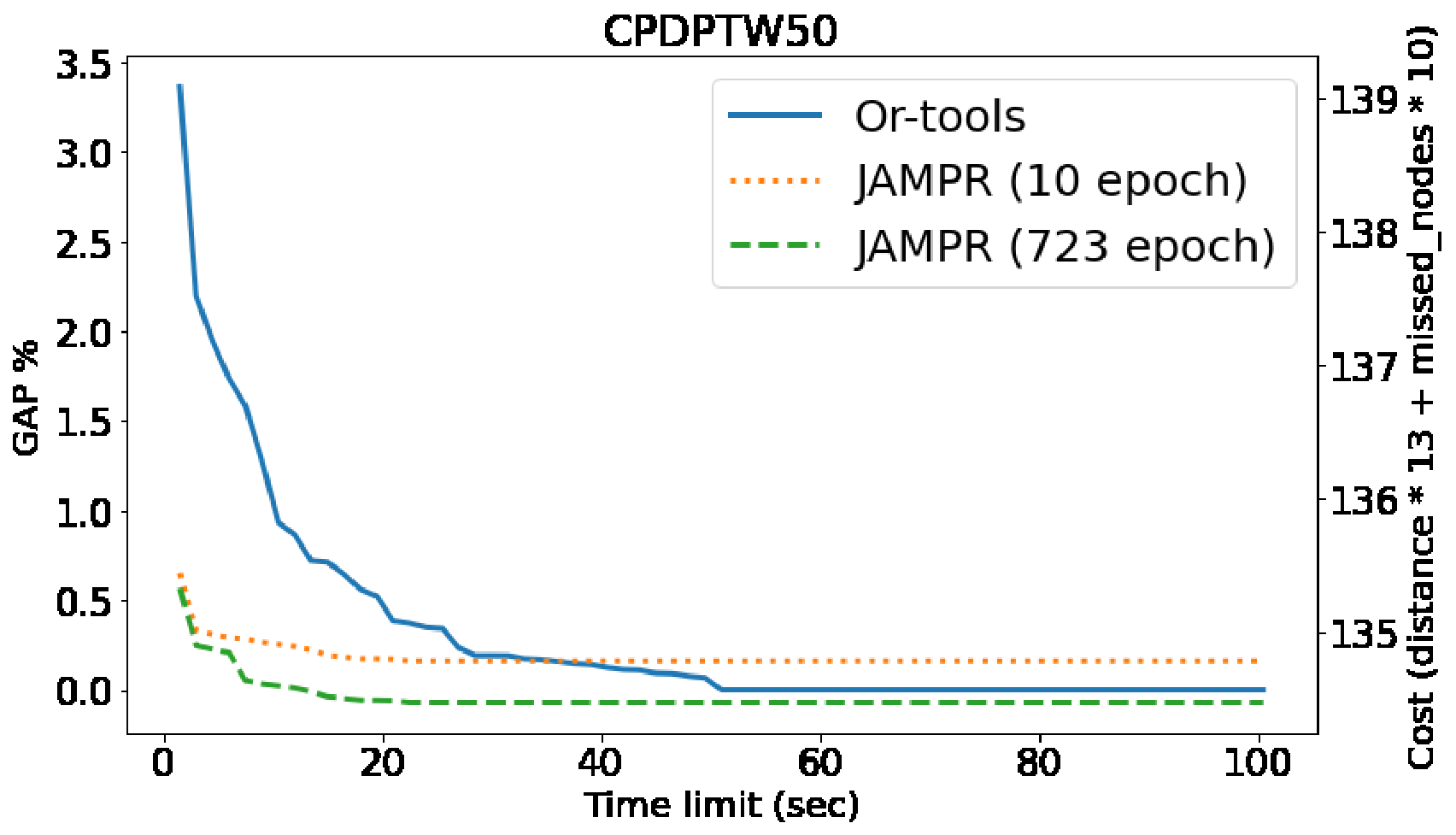}
\caption{Model performance on small-size (50) CVRPTW proplem. Dotted and dashed lines show JAMPR solution after 10 and more than 700 epochs of training, respectivelly. Well trained JAMPR model beat the heuristic solution in cost. JAMPR model after 10 epochs of taining provide fast suboptimal solution. Figure has two scales: total cost (right) and gap with respect to final baseline solution (left). }
\label{fig:first}
\end{figure}

The figure \ref{fig:first} shows graphs of changes in the quality of model prediction depending on the optimization time. The top graph shows how the cost of routes changes as the time limit for task optimization increases. The bottom plot illustrates how, as a percentage of the best OR-Tools prediction, the prediction of each model changes with increasing optimization time.

With a small dimension of the problem, the algorithm is able to give a fast optimal solution (see figure \ref{fig:first}).
Gradually, the results of the JAMPR model reach a plateau, however, due to a good approximation by the first greedy solution, the algorithm gives up to 3 percent reduction in the cost of the route in the first seconds of optimization, gradually comparing with OR-Tools. It can be seen that despite the fact that for the model that trained 723 epochs, two orders of magnitude more time was spent (160 hours versus 3), the final results, as can be seen from the graph, differ by no more than 0.3 \%. Thus, on low-dimensional problems, the impact of training time and the increase in the predictive ability of the model is disproportionate.

In addition, an important result is the number of problems for which the algorithms could not give a single solution. If JAMPR solves all problems, OR-Tools cannot solve more and more problems with increasing dimension (see figure \ref{fig:fourth}).

Thus, we can judge that for problems of small dimension, the considered algorithm is able to give the best solution in the minimum time, being preferred in the framework of CPDPTW.

\subsection{Problems of medium size}
Since the training time increases with the growth of the task size, in order to study the performance of the model on medium-sized tasks, we trained the network using the early stopping criterion: non-reduction of the path cost over 10 optimization steps. For each of the tasks, the training time is measured in days: 2, 4 days for dimensions 100 and 200, respectively. We tested each model on 100 instances of the CPDPTW problem, the p value of the metric at each point, presented on the graphs, is the average among all solved instances at a given time limit.

\begin{figure}[!h]
\centering
\includegraphics[width=0.9\textwidth]{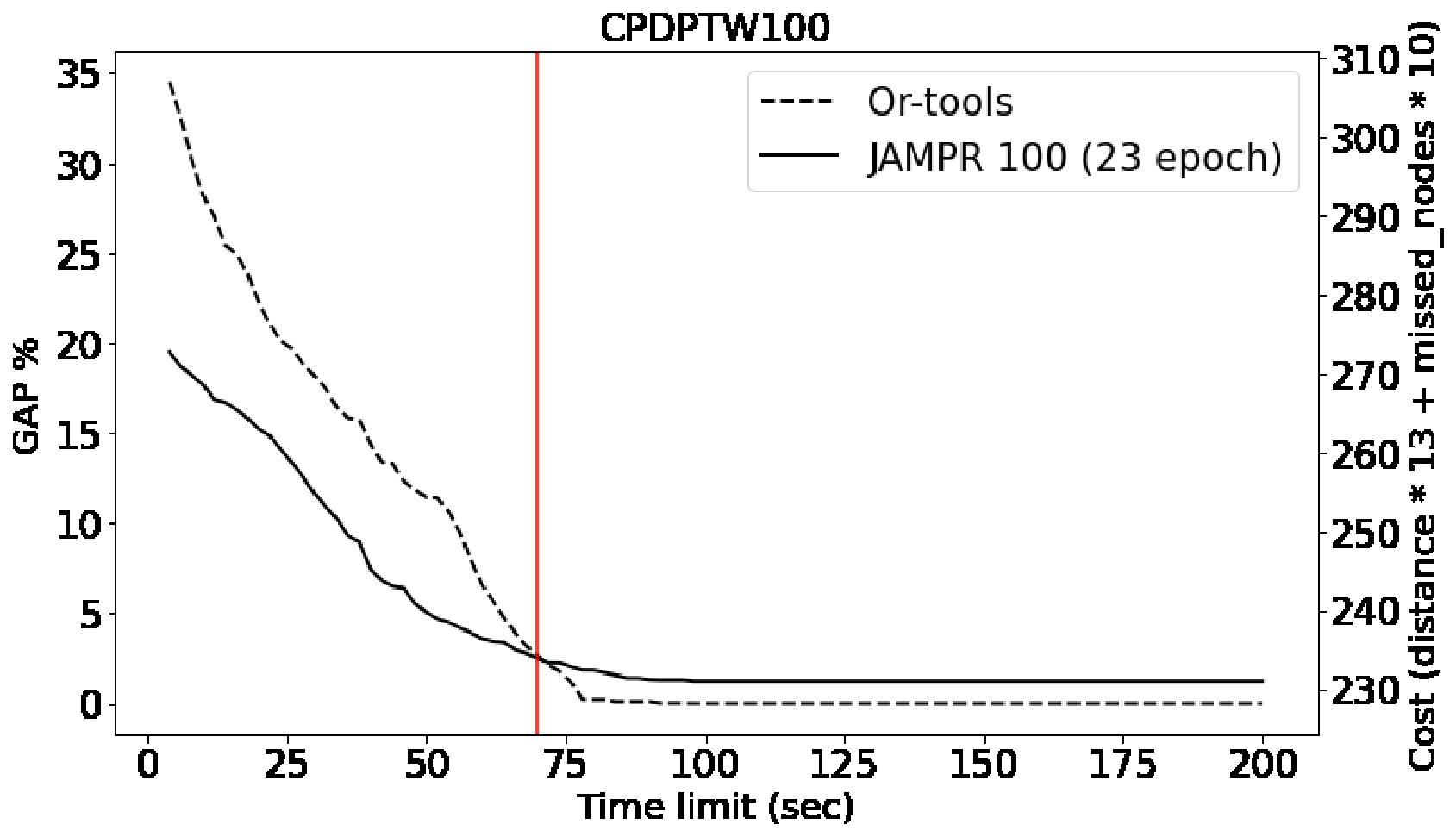}\\
\includegraphics[width=0.9\textwidth]{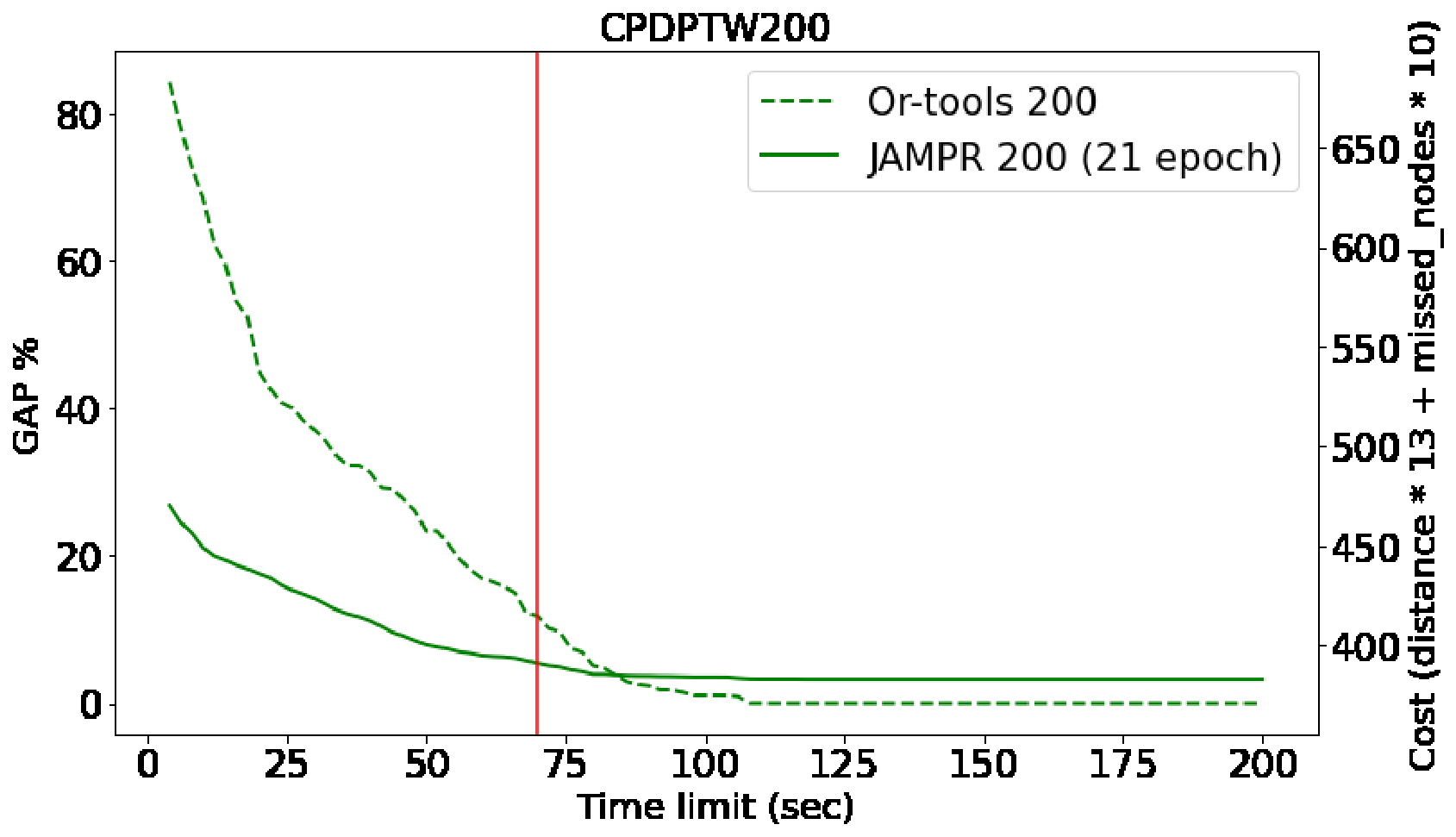}\\
\caption{Model performance on medium problem sizes (100 --- top, 200 --- bottom panel). JAMPR gives fast suboptimal result; OR-Tools and JAMPR performance are close around 70 seconds of optimization. Each figure has two scales: total cost (right) and gap with respect to final baseline solution (left).}
\label{fig:second}
\end{figure}

The figure \ref{fig:second} shows graphs of model quality changes depending on the optimization time. Each row contains different dimensions of tasks: 100 for the first, 200 for the second. The left column shows cost graphs depending on the optimization time, the right column shows the change in the percentage of the path cost depending on the optimization time. As in the case of low-dimensional problems, we considered the percentage of the cost of each model compared to the best result shown by OR-Tools for the maximum optimization time (in this case 200 seconds).

In Figure \ref{fig:second}, you can see that for more points, the algorithm exhibits similar behavior with small dimensions: a fast suboptimal solution, with a gradual decrease in optimization speed. For medium-sized problems, obviously, we cannot train the model in an adequate amount of time to get the best result. However, we can achieve a result in which the model will give a suboptimal solution in the first seconds of optimization. Despite JAMPR soon reaching a plateau, in the first seconds it outperforms metaheuristics, showing a 15 to 200 percent better result in the final path cost. As the size of the problem increases, the gap between the first fast solutions in the neural model and metaheuristics grows. These results can be useful in conditions where the route optimization run time is limited to 70 seconds.

We argue that on problems of medium dimension, the neural model is able to give a fast suboptimal solution that is superior to metaheuristics due to the large learning time. Regardless of the number of points, the behavior is the same. At the same time, as in the case of low-dimensional problems, JAMPR, in contrast to OR-Tools (see figure \ref{fig:fourth}), provides a solution for all considered problem instances.

\subsection{Large size problems}
High-dimensional problems are the most problematic due to the required time to optimize the neural model, which is sometimes critical. For the 400 points model, the early stopping occurred on the 7th day of training. To test the model we generated 100 examples of CPDPTW problem as for medium dimension problems. We set a goal to train JAMPR with a time limit of two weeks; on a task of size 1000 CPDPTW, the model managed to go through only 10 training epochs. At the same time, the model did not stop according to the principle of early stopping, described for problems of medium dimension. To test the model, we generated 10 examples of the CPDPTW problem from the distribution described in the \ref{data} paragraph, each point on the graphs is the average value of the cost of solved instances in accordance with the given optimization time limit. 

\begin{figure}[!h]
\centering
\includegraphics[width=0.9\textwidth]{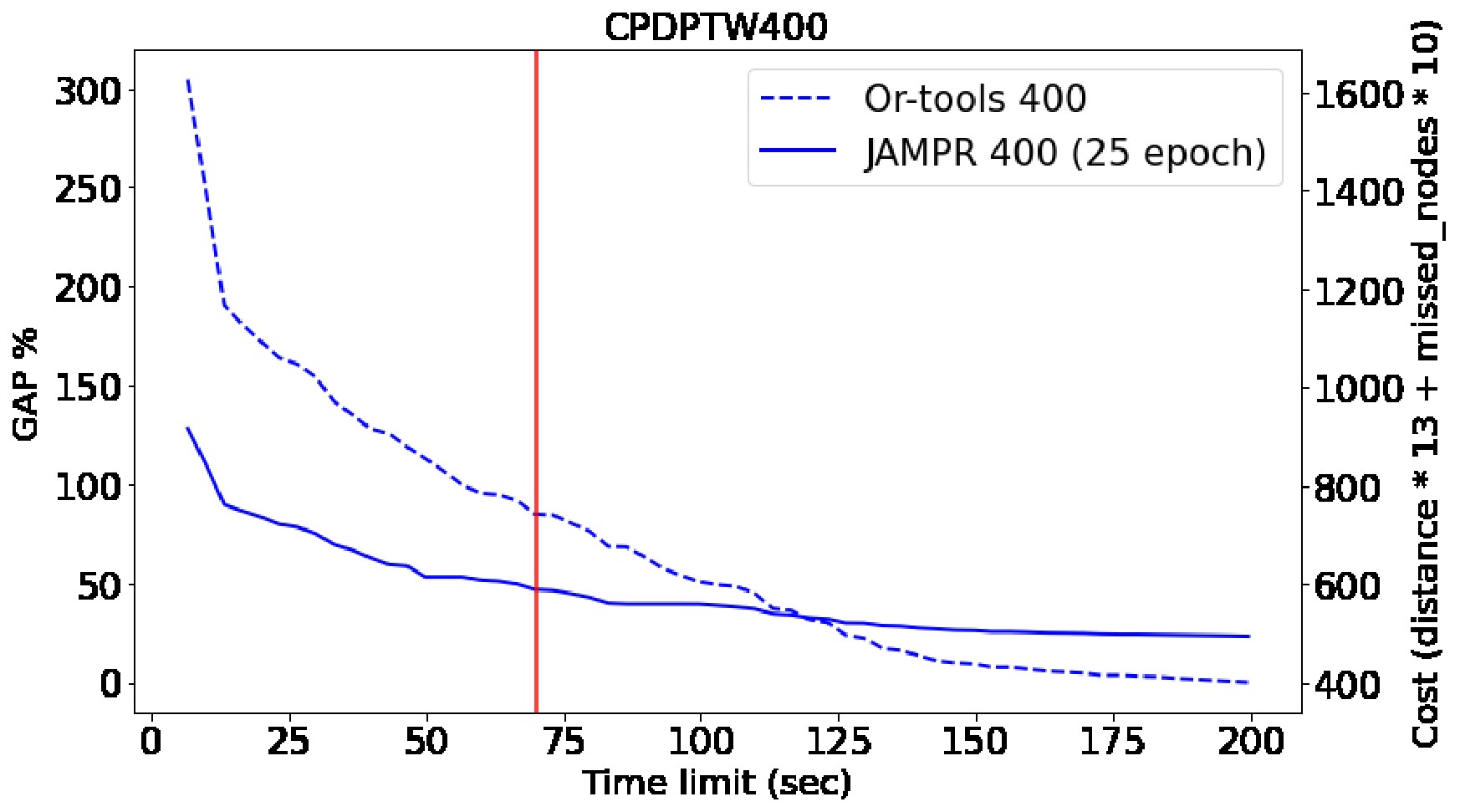}\\
\includegraphics[width=0.9\textwidth]{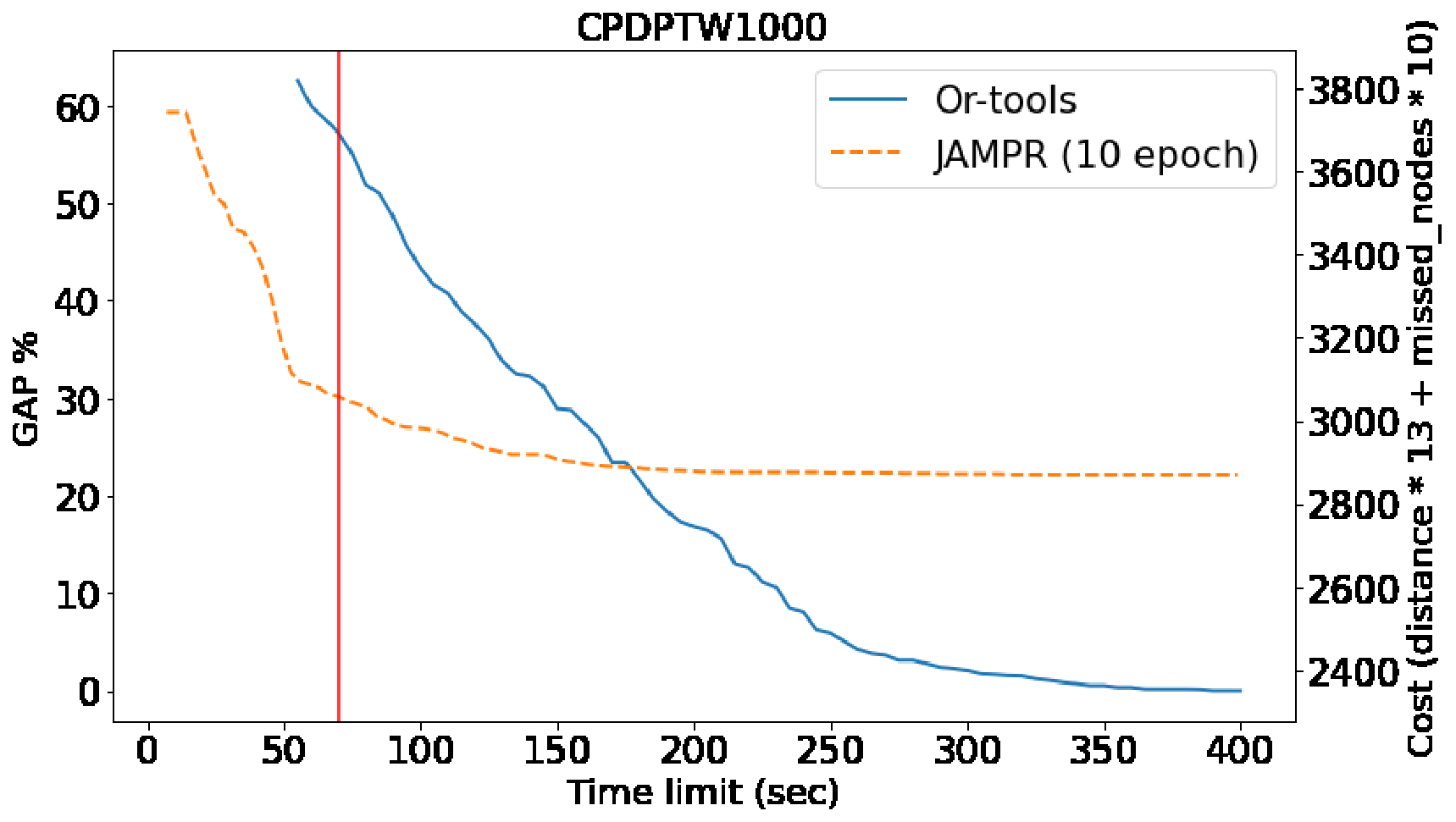}
\caption{Performance of the model at large graph sizes (400 --- top, 1000 --- bottom, respectively) looks similar to the behavior seen previously on problems of smaller size. For optimization time less than few minutes JAMPR provide a much more accurate suboptimal solution than OR-Tools baseline (note that at small optimization time OR-Tools is not able to give any solution at all). Each figure has two scales: total cost (right) and gap with respect to final baseline solution (left).}
\label{fig:third}
\end{figure}

The figure \ref{fig:third} shows how the model prediction changes depending on the time spent on task optimization. The upper part shows a graph of path cost versus optimization time, and the lower part shows a graph of the change in the percentage of path length depending on optimization time. As in the case of problems of lower dimension, we considered the percentage of the path of each model to the best result shown by OR-Tools for the maximum optimization time (in this case 400 seconds).

The behavior of the model is similar to the behavior on problems of smaller dimensions in the first seconds of optimization (see figure \ref{fig:third}): a quick solution, with a gradual plateau. 
\textcolor{black}{Over time, metaheuristics greatly overtakes the neural model. However, it takes a long time to train the model and it's impossible to improve the situation in a reasonable time.} It is worth noting that metaheuristics cannot find a single solution in the first 60 seconds of optimization, the very first proposed solution is inferior to the neural model by about 50\% of the path cost.

The results obtained on high-dimensional problems show that despite the fast suboptimal solution, neural models take too long to train. The final result of JAMPR with the maximum optimization time still loses about 20\% of the path cost to the metaheuristic. However, a fast approximate solution can be useful when it is necessary to limit the optimization time in dynamical systems.

\subsection{Analysis of model robustness}
An important property of any algorithm is its robustness. \textcolor{black}{By robustness we mean two things: 1) the model solution exists and remains stable for different instances of the CPDPTW problem in a SOFT formulation and 2) the model is able to keep the result on different distrbutions.} 

\begin{figure}[!h]
\centering
\includegraphics[width=0.95\textwidth]{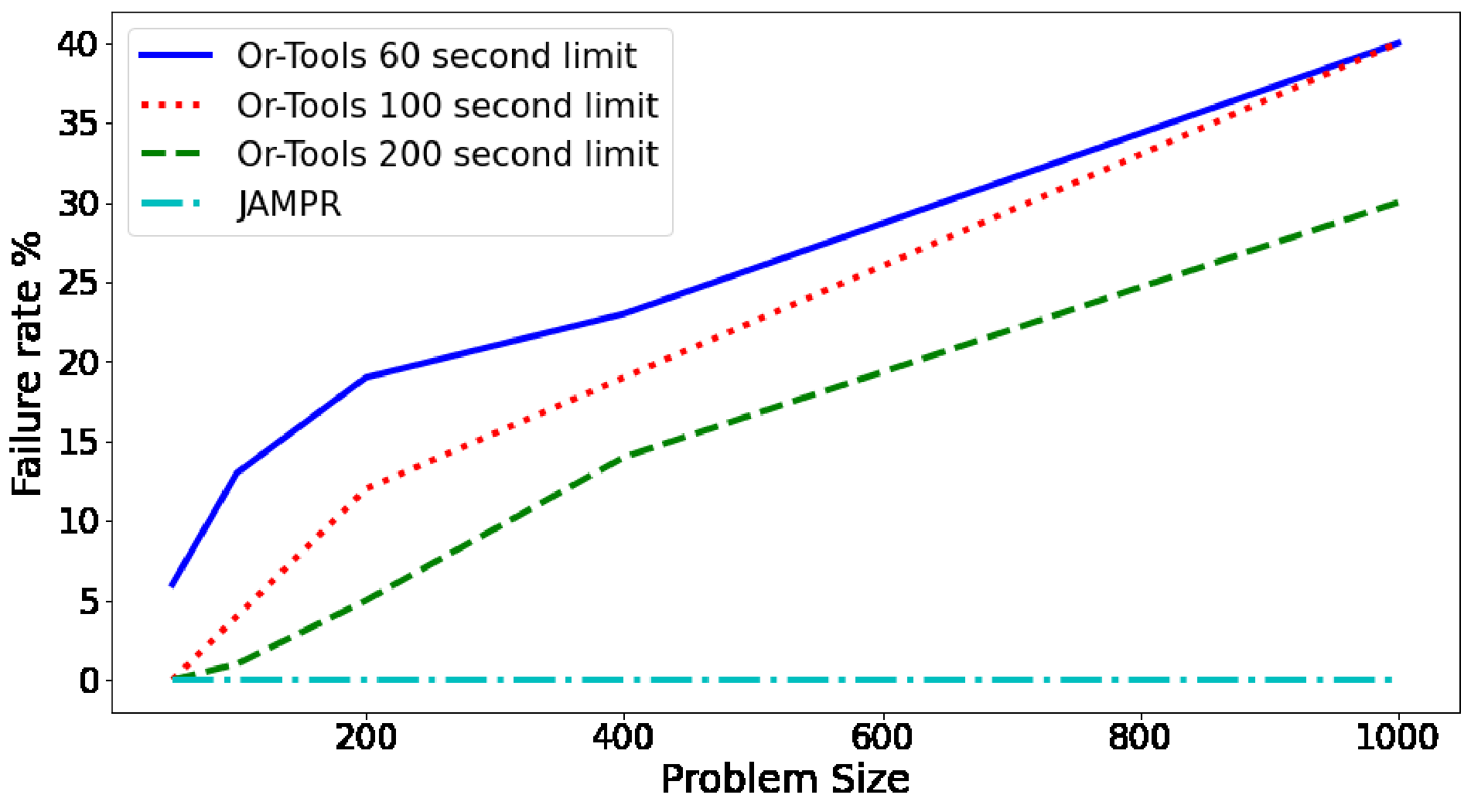}
\caption{The dependence of the number of unsolved problems on the size of the problem for different time limits. For all sizes except a thousand, 100 instances of the task were tested, for a thousand this number is 10.}
\label{fig:fourth}
\end{figure}

We noticed that regardless of the size of the problem, JAMPR was able to find a solution for all instances of the problem. With OR-Tools, the situation is different: there are 60 seconds of optimization on the horizon, the number of tasks for which a solution has not been found increases. The figure \ref{fig:fourth} shows a graph of the dependence of unsolved problems on the size of the problem. The Y-axis measures the percentage of unsolved problems to all examples of the problem, the X-axis --- the size of the problems. We considered 3 limits: 60, 100 and 200 seconds.

In the \ref{fig:fourth} figure, you can see that the JAMPR neural model was able to find solutions for all instances, while the metaheuristic increases the number of unsolved instances as the task size increases. Even on the horizon of 200 seconds, there are instances for which OR-Tools could not find a single satisfying solution.

This fact shows another important advantage of the neural RL solver --- robustness of its solution, regardless of the size of the problem.

\begin{figure}[!h]
\centering
\includegraphics[width=0.95\textwidth]{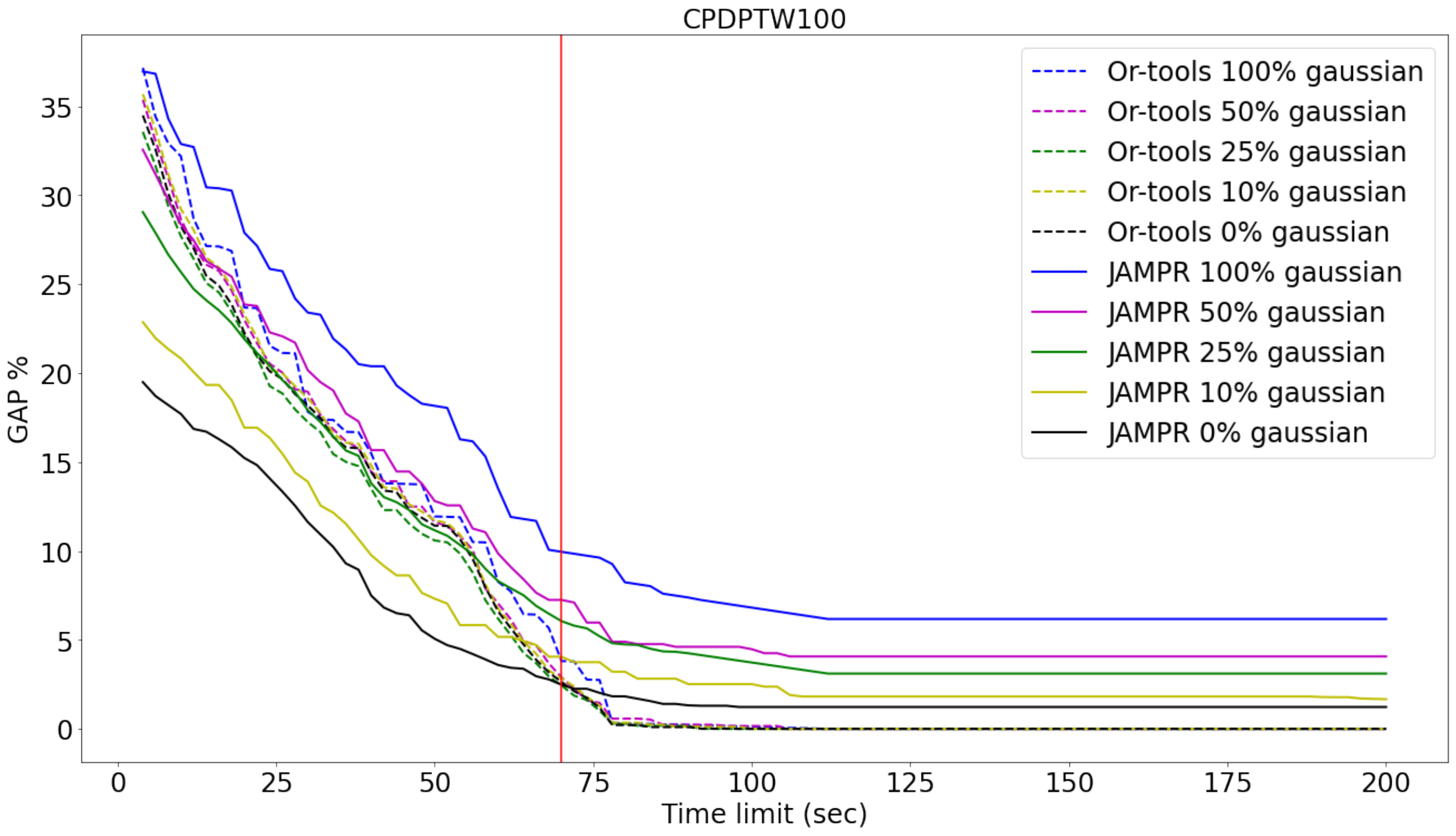}
\caption{Robustness of the model on different percentage of unknown distribution.}
\label{fig:fifth}
\end{figure}

\textcolor{black}{Another important feature of robust models is the ability to keep the result when the distribution changes. We tried to replace the uniform distribution when generating points with a normal distribution with an average at the center of the square (0.5) and a deviation of 0.1, points that go beyond the boundaries were limited by the size of the square. In the figure \ref{fig:fifth}, you can see how the absolute values of the cost of routes changed when optimizing the CPDPTW tasks when the percentage of points from the normal distribution changed from 0 to 100\%. On the y-axis --- the absolute values of the cost, on the x-axis --- the time spent on optimization.}

\textcolor{black}{We note that with an increase in the amount of data from the normal distribution, the model begins to predict worse, while OR-Tools retains its predictive ability. Nevertheless, already starting from 25\% or less of the extraneous data distribution, the model is able to represent a fast suboptimal solution that outperforms OR-Tools.}

\textcolor{black}{We additionally checked how the values of the prediction will change when the calculation of distances is changed from the Euclidean measure to the Manhattan measure: the results are exactly the same as those obtained earlier. }

\section{Conclusions}
In this work we, for the first time, successfully apply a deep Reinforcing Learning approach \textcolor{black}{(JAMPR model which we modified for additional constraints)} to solve Pickup and Delivery routing problem with Capacity and Time Window constraints (CPDPTW). We obtained a robust model that gives a fast optimal solution for problems of small and medium size (50-200), and gives fast suboptimal solution for problems of larger size.

\textcolor{black}{We note:
\begin{itemize}
    \item For all problem size, using the JAMPR model, we can get a fast suboptimal solution in the first tenths of seconds. For small and medium size (50-200) problems a final JAMPR solution is very close to OR-Tools baseline with lower optimization time. With more training time JAMPR performance can be improved even more, but for large size problems it takes too much computational time on one GPU.
    \item The neural model is more robust than OR-Tools metaheuristics for CPDPTW tasks in terms of existence of solution for all instances for the same distribution. JAMPR has zero failure rate even for small optimization time and large size problems; the OR-Tools metaheuristics failure rate (for the fixed optimization time) increases with problem size.
    \item The JAMPR model is robust for a changing of distribution. It saves the behavior of a fast suboptimal solution in the presence of up to $\sim25\%$ of data from normal radially symmetrical distribution.
\end{itemize}}

We conclude that deep reinforcement learning algorithms, like JAMPR model, can be considered as promising model for solving high-dimensional route optimization problems with time-window and capacity constraints.

\section{Future work}

What is the best solver for low-dimensional subproblems? To the best of our knowledge, optimization of routes using heuristics with a trained reinforcement model gives SotA results in this area, like Lu et al. \cite{lu2019learning}. These approaches show the best results for the low-dimensional problems, but are difficult to implement in terms of creating rules to support new constraints, nevertheless it remains an interesting area of research.

We also want to note that the use of mixed-integer programming algorithms gives the exact solution for low-dimensional problems, but is not applicable even for medium-sized problems due to the impressive time required for optimization. We consider it an interesting direction to study: the combination of reinforcement learning algorithms and MIP solvers for solving CPDPTW problems.

It is well known that solving high-dimensional tasks is a challenging problem and requires special approaches. One of the most popular solutions is splitting the problem into subtasks and solving these small subtasks using classical methods. At the moment, the best results are shown by the learned partitioning policies based on neural networks. For example, Li et al. \cite{li2021learning}, uses a Transformer \cite{vaswani2017attention} architecture to train a classifier/regressor, showing how correct the choice of a specific subproblem for re-optimization is. We plan to combine this approach with the kind of JAMPR model considered here in order to achieve a better performance and training speed up for the high-dimensional problems.

%
%
%
\bibliographystyle{splncs04}
\bibliography{mybibliography}

@article{lin1973effective,
  title={An effective heuristic algorithm for the traveling-salesman problem},
  author={Lin, Shen and Kernighan, Brian W},
  journal={Operations research},
  volume={21},
  number={2},
  pages={498--516},
  year={1973},
  publisher={Informs}
}

@article{clarke1964scheduling,
  title={Scheduling of vehicles from a central depot to a number of delivery points},
  author={Clarke, Geoff and Wright, John W},
  journal={Operations research},
  volume={12},
  number={4},
  pages={568--581},
  year={1964},
  publisher={Informs}
}

@article{braysy2005vehicle,
  title={Vehicle routing problem with time windows, Part I: Route construction and local search algorithms},
  author={Br{\"a}ysy, Olli and Gendreau, Michel},
  journal={Transportation science},
  volume={39},
  number={1},
  pages={104--118},
  year={2005},
  publisher={INFORMS}
}

@inproceedings{perron2011operations,
  title={Operations research and constraint programming at google},
  author={Perron, Laurent},
  booktitle={International Conference on Principles and Practice of Constraint Programming},
  pages={2--2},
  year={2011},
  organization={Springer}
}

@article{vidal2012hybrid,
  title={A hybrid genetic algorithm for multidepot and periodic vehicle routing problems},
  author={Vidal, Thibaut and Crainic, Teodor Gabriel and Gendreau, Michel and Lahrichi, Nadia and Rei, Walter},
  journal={Operations Research},
  volume={60},
  number={3},
  pages={611--624},
  year={2012},
  publisher={INFORMS}
}

@article{vidal2022hybrid,
  title={Hybrid genetic search for the CVRP: Open-source implementation and SWAP* Neighborhood},
  author={Vidal, Thibaut},
  journal={Computers \& Operations Research},
  volume={140},
  pages={105643},
  year={2022},
  publisher={Elsevier}
}

@article{li2021learning,
  title={Learning to delegate for large-scale vehicle routing},
  author={Li, Sirui and Yan, Zhongxia and Wu, Cathy},
  journal={Advances in Neural Information Processing Systems},
  volume={34},
  year={2021}
}

@article{vaswani2017attention,
  title={Attention is all you need},
  author={Vaswani, Ashish and Shazeer, Noam and Parmar, Niki and Uszkoreit, Jakob and Jones, Llion and Gomez, Aidan N and Kaiser, {\L}ukasz and Polosukhin, Illia},
  journal={Advances in neural information processing systems},
  volume={30},
  year={2017}
}

@article{falkner2020learning,
  title={Learning to solve vehicle routing problems with time windows through joint attention},
  author={Falkner, Jonas K and Schmidt-Thieme, Lars},
  journal={arXiv preprint arXiv:2006.09100},
  year={2020}
}

@article{kool2018attention,
  title={Attention, learn to solve routing problems!},
  author={Kool, Wouter and Van Hoof, Herke and Welling, Max},
  journal={arXiv preprint arXiv:1803.08475},
  year={2018}
}

@article{solomon1987algorithms,
  title={Algorithms for the vehicle routing and scheduling problems with time window constraints},
  author={Solomon, Marius M},
  journal={Operations research},
  volume={35},
  number={2},
  pages={254--265},
  year={1987},
  publisher={Informs}
}

@inproceedings{lu2019learning,
  title={A learning-based iterative method for solving vehicle routing problems},
  author={Lu, Hao and Zhang, Xingwen and Yang, Shuang},
  booktitle={International conference on learning representations},
  year={2019}
}

@article{nazari2018reinforcement,
  title={Reinforcement learning for solving the vehicle routing problem},
  author={Nazari, Mohammadreza and Oroojlooy, Afshin and Snyder, Lawrence and Tak{\'a}c, Martin},
  journal={Advances in neural information processing systems},
  volume={31},
  year={2018}
}

@article{vinyals2015pointer,
  title={Pointer networks},
  author={Vinyals, Oriol and Fortunato, Meire and Jaitly, Navdeep},
  journal={Advances in neural information processing systems},
  volume={28},
  year={2015}
}

@article{chen2019learning,
  title={Learning to perform local rewriting for combinatorial optimization},
  author={Chen, Xinyun and Tian, Yuandong},
  journal={Advances in Neural Information Processing Systems},
  volume={32},
  year={2019}
}

@article{savelsbergh1992vehicle,
  title={The vehicle routing problem with time windows: Minimizing route duration},
  author={Savelsbergh, Martin WP},
  journal={ORSA journal on computing},
  volume={4},
  number={2},
  pages={146--154},
  year={1992},
  publisher={INFORMS}
}

@book{or1976traveling,
  title={TRAVELING SALESMAN TYPE COMBINATORIAL PROBLEMS AND THEIR RELATION TO THE LOGISTICS OF REGIONAL BLOOD BANKING.},
  author={Or, Ilhan},
  year={1976},
  publisher={Northwestern University}
}

@article{parragh2008survey,
  title={A survey on pickup and delivery problems},
  author={Parragh, Sophie N and Doerner, Karl F and Hartl, Richard F},
  journal={Journal f{\"u}r Betriebswirtschaft},
  volume={58},
  number={1},
  pages={21--51},
  year={2008},
  publisher={Springer}
}

@book{wolsey1999integer,
  title={Integer and combinatorial optimization},
  author={Wolsey, Laurence A and Nemhauser, George L},
  volume={55},
  year={1999},
  publisher={John Wiley \& Sons}
}

@article{dantzig1954solution,
  title={Solution of a large-scale traveling-salesman problem},
  author={Dantzig, George and Fulkerson, Ray and Johnson, Selmer},
  journal={Journal of the operations research society of America},
  volume={2},
  number={4},
  pages={393--410},
  year={1954},
  publisher={INFORMS}
}

@article{little1963algorithm,
  title={An algorithm for the traveling salesman problem},
  author={Little, John DC and Murty, Katta G and Sweeney, Dura W and Karel, Caroline},
  journal={Operations research},
  volume={11},
  number={6},
  pages={972--989},
  year={1963},
  publisher={INFORMS}
}
%




\end{document}